\documentclass[runningheads]{llncs}

\usepackage{subcaption}
\usepackage{marvosym}
\usepackage{array}
\usepackage[T1]{fontenc}
\usepackage{amssymb}
\usepackage{amsmath}
\usepackage{graphicx}
\usepackage{tabularx}
\usepackage{pict2e}
 \usepackage{colortbl}
\usepackage{tikz}

\usepackage{mathrsfs}
\usepackage{bbm}

\usepackage{booktabs}
\usepackage{multirow}
\usepackage{multicol}
\usepackage{threeparttable}

\usepackage{caption}
\usepackage{subfig}
\usepackage{stackengine}
\usepackage{makecell}

\usepackage{color}

\def\doi#1{\href{https://doi.org/\detokenize{#1}}{\url{https://doi.org/\detokenize{#1}}}}
\graphicspath{{LaTeX2e+Proceedings+Templates+download/}}

\begin{document}
\title{Multi-Agent Reasoning for Cardiovascular Imaging Phenotype Analysis}

\titlerunning{MESHAgents}
\authorrunning{Zhang W., Qiao M. et al.}


\author{Weitong Zhang\inst{1}$^{*}$, Mengyun Qiao\inst{2}$^{*}$, Chengqi Zang\inst{5}, Steven Niederer\inst{6}, Paul M Matthews\inst{3,8,9}, 
Wenjia Bai\inst{1,3,4}, Bernhard Kainz\inst{1,7}}
\institute{Department of Computing, Imperial College London, London, UK
\and Department of Mechanical Engineering, University College London, London, UK
\and Department of Brain Sciences, Imperial College London, London, UK 
\and Data Science Institute, Imperial College London, London, UK
\and University of Tokyo, Tokyo, JP
\and National Heart and Lung Institute, Imperial College London, London, UK
\and FAU Erlangen-Nürnberg, Erlangen, DE
\and UK Dementia Research Institute, Imperial College London, London, UK
\and Rosalind Franklin Institute, Harwell Science and Innovation Campus, Didcot, UK
\\
\email{wz1820@ic.ac.uk,m.qiao@ucl.ac.uk}
}
\maketitle
\let\thefootnote\relax\footnotetext{$^{*}$ Equal contribution}

\begin{abstract}
Identifying associations between imaging phenotypes, disease risk factors, and clinical outcomes is essential for understanding disease mechanisms.
However, traditional approaches rely on human-driven hypothesis testing and selection of association factors, often overlooking complex, non-linear dependencies among imaging phenotypes and other multi-modal data. To address this, we introduce\textbf{M}ulti-agent \textbf{E}xploratory \textbf{S}ynergy for the \textbf{H}eart (MESHAgents): a  framework that leverages large language models as agents to dynamically elicit, surface, and decide confounders and phenotypes in association studies.
Specifically, we orchestrate a multi-disciplinary team of AI agents, which spontaneously generate and converge on insights through iterative, self-organizing reasoning. The framework dynamically synthesizes statistical correlations with multi-expert consensus, providing an automated pipeline for phenome-wide association studies (PheWAS).
We demonstrate the system's capabilities through a population-based study of imaging phenotypes of the heart and aorta. 
MESHAgents autonomously uncovered correlations between imaging phenotypes and a wide range of non-imaging factors, identifying additional confounder variables beyond standard demographic factors. Validation on diagnosis tasks reveals that MESHAgents-discovered phenotypes achieve performance comparable to expert-selected phenotypes, with mean AUC differences as small as $-0.004_{\pm0.010}$ on disease classification tasks. Notably, the recall score improves for 6 out of 9 disease types. Our framework provides clinically relevant imaging phenotypes with transparent reasoning, offering a scalable alternative to expert-driven methods.

\end{abstract}
\section{Introduction}

Medical imaging is essential for deriving quantitative phenotypes of anatomical structure and function and understanding their associations with various non-imaging factors, such as genetic predispositions~\cite{sing2003genes}, environmental conditions~\cite{cosselman2015environmental} and physiological characteristics~\cite{rozanski1999impact}.
How to better integrate imaging with non-imaging data to perform multi-modal data analysis has received increasing attention in the imaging community. To ensure reliable and unbiased associations between imaging and non-imaging data, it is important to identify associations affecting both risk factors and imaging phenotypes~\cite{skelly2012assessing}. A confounder is an association variable that influences both the independent variable and the dependent variable, creating a spurious association between them. Ignoring confounders can lead to misinterpretations in phenome-wide association studies (PheWAS) and genome-wide association studies (GWAS), impacting clinical analyses. Confounding factors also need
to be considered when evaluating bias in machine learning algorithms~\cite{mukherjee2022confounding}, particularly in medical imaging research~\cite{qiao2023cheart,qiao2024personalised}. Conventional methods rely on established empirical criteria or expert selection, not accounting for complicated non-linear relationships between imaging phenotypes and health outcomes. The automated integration of reasoning across large-scale, multi-modal datasets and diverse expert domains remains an underexplored area.

Recent advances in large language models (LLMs)~\cite{touvron2023llama,openai2023gpt4} have exhibited notable reasoning abilities across a wide range of tasks and applications~\cite{lu2023chameleon,Park2023GenerativeAgents}, with these capabilities stemming from extensive training on vast comprehensive corpora covering diverse topics. In real-world scenarios, LLMs tend to encounter domain-specific tasks (\textit{e.g.}, strategy~\cite{zhang2024BiD}, safety~\cite{zhang2024truthdeceitbayesiandecoding}) that necessitate a combination of domain expertise and complex reasoning abilities~\cite{moor2023foundation,clinical}. This backdrop makes the adoption of LLMs in the medical field a noteworthy research topic, which has recently gained growing attention~\cite{zhang2023alpacareinstructiontuned,bao2023discmedllm}.

\begin{figure}[t]
    \centering
    \resizebox{0.7\columnwidth}{!}{%
    \begin{minipage}[b]{0.4\textwidth}
        \centering
        \scriptsize
        \resizebox{\linewidth}{!}{\begin{tabular}{l|l}
        \toprule
        \textbf{Expert-Selected\cite{bai2020population,qiao2024personalised} } & \textbf{MESHAgents-Discovered} \\
        \midrule
        \multicolumn{2}{c}{\textbf{Imaging phenotypes}} \\
        \midrule
        RVEDV (mL) \cellcolor{green!15} & RVEDV (mL) \cellcolor{green!15} \\
        LVEDV (mL) & RVESV (mL) \\
        LVEF (\%) & RVSV (mL) \\
        LVM (g) & LVCO (L/min) \\
        RVEF (\%) & LAV min (mL) \\
        LAV max (mL) & LASV (mL) \\
        LAEF (\%) & RAV max (mL) \cellcolor{green!15} \\
        RAV max (mL) \cellcolor{green!15} & RAV min (mL) \\
        RAEF (\%) & AAo max area (mm2) \cellcolor{green!15} \\
        AAo max area (mm2) \cellcolor{green!15} & AAo min area (mm2) \\
        AAo distensibility & DAo min area (mm2) \\
        DAo max area (mm2) & WT\_AHA\_2 (mm) \\
        DAo distensibility & Ell\_4 (\%) \\
        \midrule
        \multicolumn{2}{c}{\textbf{Confounders}} \\
        \midrule
        Sex \cellcolor{green!15} & Sex \cellcolor{green!15} \\
        Age \cellcolor{green!15} & Age \cellcolor{green!15} \\
        Weight \cellcolor{green!15} & Weight \cellcolor{green!15} \\
        Height \cellcolor{green!15} & Height \cellcolor{green!15} \\
        & Alcohol intake \\
        & Systolic BP \\
        & Diastolic BP \\
        \bottomrule
        \end{tabular}
        }
        \caption*{(a) Parameter comparison}
    \end{minipage}
    \hfill
    \begin{minipage}[b]{0.59\textwidth}
        \centering
        \includegraphics[width=\textwidth]{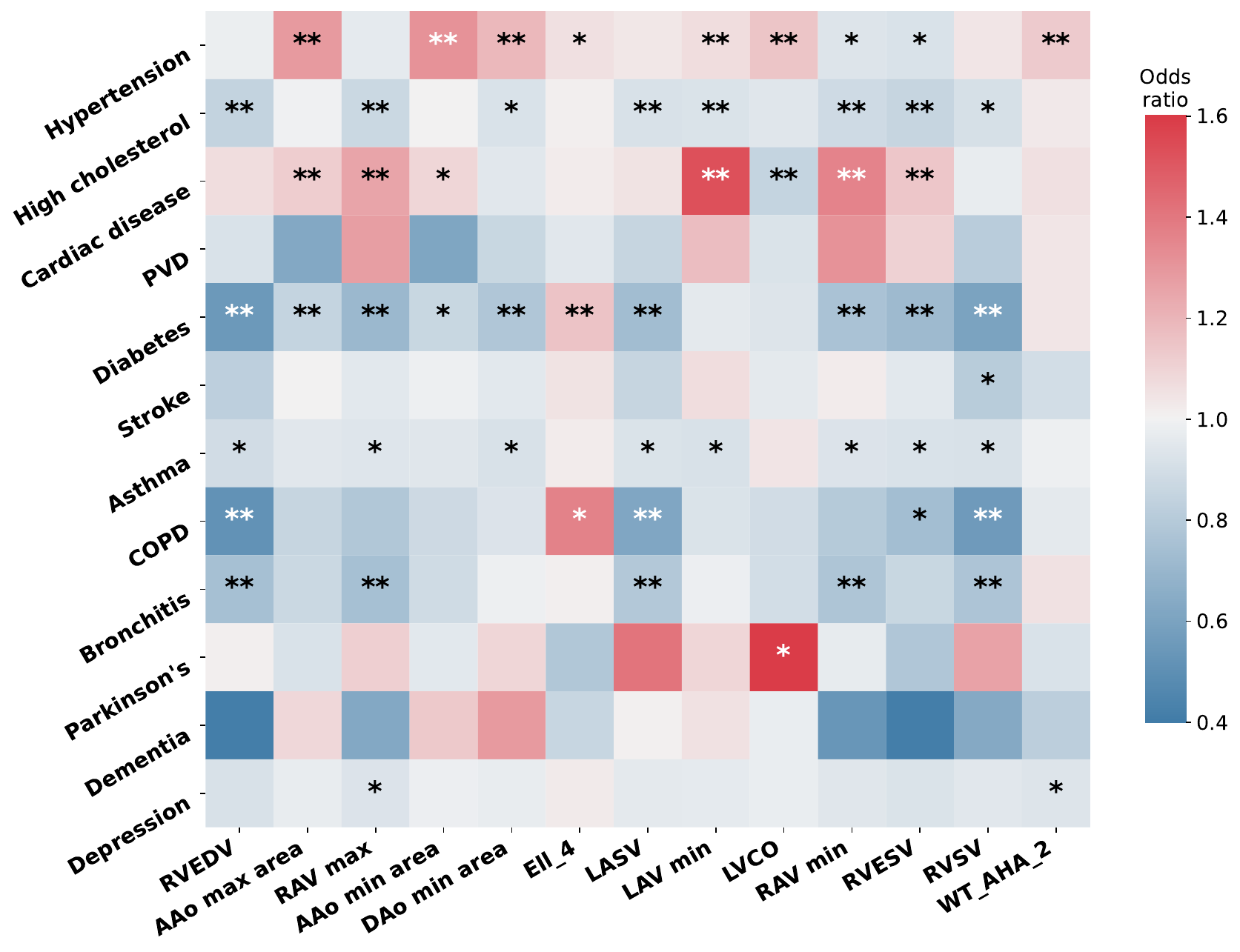}
        \caption*{(b) Disease association heatmaps}
    \end{minipage}
    }
    \caption{\textbf{MESHAgents-discovered cardiovascular parameters and their disease associations.} (a) Comparison between expert-selected in literature~\cite{bai2020population,qiao2024personalised} and MESHAgents-discovered parameters, with overlapping parameters highlighted in green. MESHAgents autonomously identified clinically relevant phenotypes and additional confounders without prior domain knowledge. (b) Heatmaps showing significant associations between discovered imaging phenotypes and disease outcomes on 38,309 participants (**p < 0.01, *p < 0.05).}
    \label{fig:result1}
\end{figure}

\noindent\textbf{Challenges.} Two major challenges prevent LLMs from handling tasks in multi-modal medical imaging data analysis:
(i) Imaging phenotype analysis demands {multi-disciplinary domain expertise}~\cite{lievin2022can,schmidt2007expertise} spanning medicine, biomechanics, and statistics, yet monolytic LLMs suffer from {catastrophic forgetting}~\cite{taraledistributed,zheng2024towards} in maintaining comprehensive knowledge across disciplines. 
(ii) The analysis requires {phenotype-aware reasoning} to identify robust correlations between imaging phenotypes and a wide range of non-imaging factors~\cite{skelly2012assessing,bai2020population}, whereas LLMs tend to produce \textit{hallucinations}~\cite{harris2023large,ji2023survey} that undermine phenotype discovery.

\noindent\textbf{Contributions.} 
(i) 
We introduce the first multi-agent system for medical imaging PheWAS where domain-specific LLM agents achieve association consensus through structured interactions, mitigating catastrophic forgetting through a multi-disciplinary team.
(ii) 
We demonstrate that role-based collaborative reasoning enhances imaging phenotype discovery without retrieval augmentation, maintaining interpretable decision trails while reducing hallucination.
(iii) 
Through multi-faceted evaluation, MESHAgents shows superior phenotype discovery over existing single-agent and multi-agent approaches on population imaging data\footnote{We follow the evaluation setting from literature~\cite{bai2020population}.}, while achieving diagnostic performance comparable to expert-selected parameters across three classification models (AUC difference $-0.004{_{\pm0.010}}$)

\noindent\textbf{Related Work.} Recently, medical AI agents have shown promising progress, from single-agent systems like Med-PaLM~\cite{singhal2023large} to collaborative frameworks such as MedAgents~\cite{tang2023medagents} and RareAgents~\cite{chen2024rareagents} for medical question answering and diagnosis. These systems have demonstrated the potential of LLM-based agents in clinical scenarios, including triaging~\cite{lu2024triageagent} and medical image analysis~\cite{li2024mmedagent}. Meanwhile, multi-agent collaboration has emerged as a powerful paradigm in LLM applications~\cite{li2023metaagents}, with approaches like role playing~\cite{wang2023unleashing} and structured communication~\cite{wu2023autogen} enabling more sophisticated reasoning. Recent works have explored adversarial collaboration through debates~\cite{du2023improving} and negotiation~\cite{fu2023improving} to enhance decision quality. However, these approaches lack consensus mechanisms for orchestrating distributed medical expertise and validating phenotype associations, making them insufficient for faithful medical imaging PheWAS.

\section{\textbf{M}ulti-agent \textbf{E}xploratory \textbf{S}ynergy for the \textbf{H}eart}
\begin{figure}[t] 
    \centering
    \includegraphics[width=0.95\linewidth]{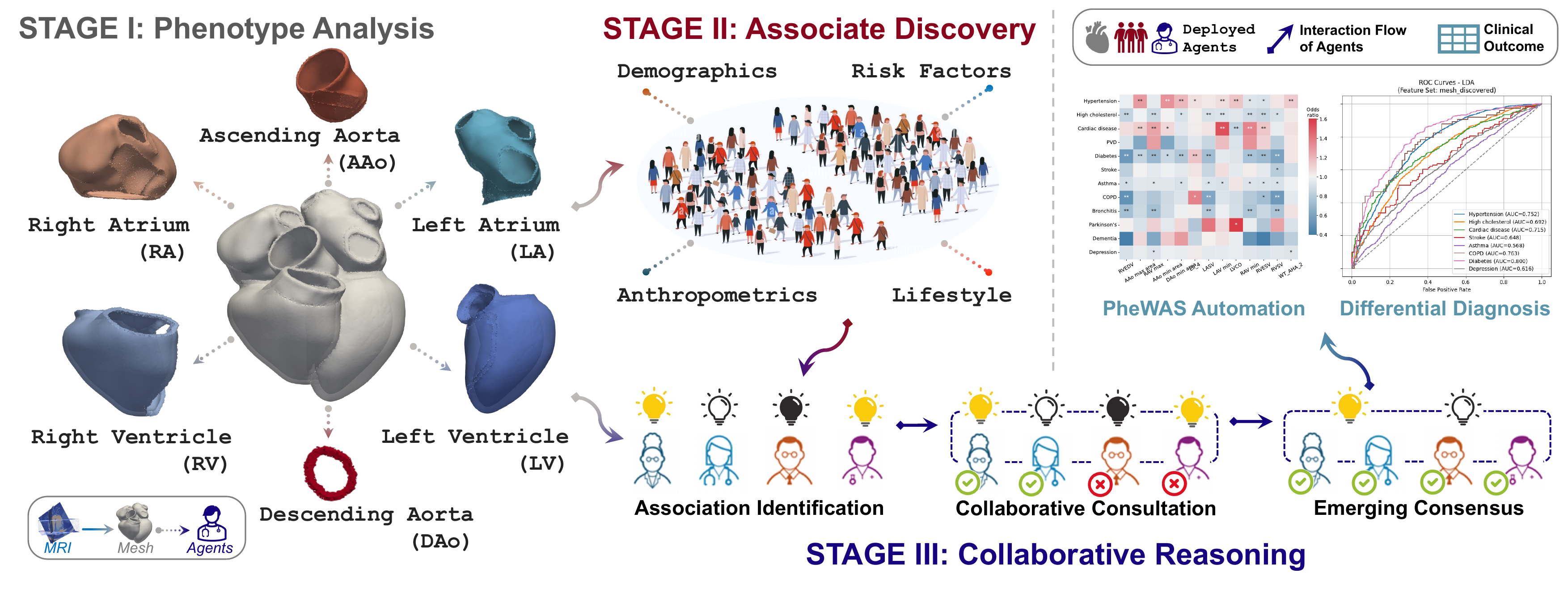}
    \caption{\textbf{\texttt{MESHAgents} framework for cardiac imaging phenotype analysis through multi-agent reasoning.} Operating and validating in three stages: (i) specialist agents analyze cardiac structures from imaging measurements; (ii) agents discover confounders via distributed analysis; (iii) collaborative reasoning generates phenome-wide associations, validated through clinical metrics.}
    \label{fig:main}
\end{figure}

\noindent\textbf{Problem Formulation.}
(i) For phenotype association analysis, 
given a large-scale imaging dataset containing $N$ participants, let $\mathcal{X} = \{x_1, ..., x_N\}$ represent the set of imaging-derived phenotypes where each $x_i \in \mathbb{R}^d$ consists of $d$ structural and functional phenotypes. For each participant, we also observe a set of non-imaging factors $\mathcal{C} = \{c_1, c_2, ..., c_M\}$ spanning demographics, anthropometrics, lifestyle, and risk factors. Our objective is to discover a set of phenotypes $\mathcal{P} = \{p_1, p_2, ..., p_K\}$ where each phenotype $p_k$ exhibits a strong and meaningful association with certain factors. MESHAgents identifies phenome-wide associations $\mathcal{A} = \{(p_k, c_m, a_{km})\}$ between image phenotype $p_k$ and factor $c_m$. 
(ii) For diagnosis, we apply MESHAgents-discovered features to a disease classification model. Each patient profile is represented using MESHAgents-identified parameters, including both imaging phenotypes and non-imaging factors: $\mathcal{X}_i^{MESH} = \{p_k(x_i), c_m\}$ where $p_k \in \mathcal{P}$ and $c_m \in \mathcal{C}$. Standard classifiers then classify these features to a disease type: $f_{classifier}(\mathcal{X}_i^{MESH}) \rightarrow \mathcal{D} = \{d_1, d_2,..., d_9\}$. This provides a more stringent evaluation than query-based tasks~\cite{chen2024rareagents}, though MESHAgents could support direct diagnostic querying with information retrieval.

\noindent\textbf{MESHAgents Overview.} MESHAgents orchestrates an expert team of agents $\alpha_i$ to analyze imaging phenotypes and discover associative factors through three progressive stages: $\mathcal{R}$ is medical imaging data containing structural and functional measurements, and $\mathcal{A}$ represent the identified phenotype-factor relationships contributing to diagnosis.

\noindent\textbf{Dynamic Memory.} To enable collective analysis, we design memory-augmented agents that integrates domain expertise with historical experience across local analysis and group reasoning. Each specialist agent $\alpha_i$ maintains a structured, long-term memory bank $\mathcal{M}_i$ that encodes imaging phenotype patterns and associative factors from past analyses. For a given phenotype set $\mathcal{X}$, the framework employs a similarity-based mechanism:
$\mathcal{R}_i = \arg\max_{h \in \mathcal{H}_i} \text{Sim}$ $(Emb(\mathcal{X}), Emb(h))$, where $\mathcal{H}_i$ represents agent $\alpha_i$'s historical cases and $\mathcal{R}_i$ is the retrieved relevant experience. $Emb(\cdot)$ maps imaging phenotypes to an embedding space. This memory mechanism enables agents to leverage past experiences and statistical evidence when evaluating phenotypes $\mathcal{P}$ and factors $\mathcal{C}$.

\noindent\textbf{Consensus Mechanism.} To foster independent reasoning while preventing premature convergence, MESHAgents employ a sequential discussion protocol. In each round $t$, expert agent $\alpha_i$ formulates its opinion independently based on three essential components:
$O_i^t = g_i(\mathcal{V}_i, \mathcal{H}_{t-1}, \mathcal{M}_i)$, where $g_i(\cdot)$ is the opinion generation of agent $\alpha_i$ that synthesizes domain-specific insights, discussion history, and group context. Domain analysis $\mathcal{V}_i$ ensures specialty-specific insights, discussion history $\mathcal{H}_{t-1}$ enables progressive refinement, and memory bank $\mathcal{M}_i$ provides group context. This sequential consensus building offers advantages over simultaneous discussion: (i) it reduces mutual influence during initial assessment, allowing each expert to form an unbiased domain-specific opinion~\cite{yang2024oasis}, (ii) it enables integration and update of previous insights through $\mathcal{H}_{t-1}$ while avoiding false consensus~\cite{zhang2024truthdeceitbayesiandecoding}, and (iii) it provides transparent decision trails instead of black-box models. The coordinator agent guides this process until reaching agreement (typically within 10 rounds), ensuring both independent expertise contribution and coherent consensus formation. Critically, this sequential mechanism creates traceable emergence patterns and transparent validation, whether phenotype analysis, associative factor discovery, or collective reasoning process.

\noindent\textbf{Tools.}
Each agent use a suite of tools $\mathcal{T} = \{T_1, ..., T_J\}$ that process imaging phenotypes $\mathcal{R}$. Each tool $T_j$ implements specific analytical functions:
$\mathcal{T}\mathcal{R} = \bigcup_{j=1}^J T_j(\mathcal{R})$, where this represents the collective evidence generated by applying agent-selected tools to the data.
These tools perform statistical significance tests, effect size calculation, and population distribution analysis. The final decision $\mathcal{A}$ emerges through synthesis of agent opinions $\mathcal{O}^{(t)}$ at discussion round $t$, memory retrievals $\mathcal{R}_i$, and tool-derived evidence $\mathcal{T}\mathcal{R}$:
$\mathcal{A} = f_{AP}(\{\mathcal{O}^{(t)}\}_{t=1}^T$, $\{\mathcal{R}_i\}_{i=1}^N, \mathcal{T}\mathcal{R})$, where $f_{AP}$ represents an aggregation function that weights different information sources based on their statistical confidence (< 0.05) and on-topic relevance (> 0.3). This integrated framework provides the foundation for the three-stage analysis pipeline, enabling phenotype valuation $\mathcal{V}_i$, factor effect size assessment $\mathcal{E}_i$, and collaborative reasoning across three stages:

\texttt{Stage I: Phenotype Analysis.}
In the initial stage, each agent $\alpha_i$ analyses the imaging phenotypes within a specific domain: \{left ventricle (LV), right ventricle (RV), left atrium (LA), right atrium (RA), ascending aorta (AAo) and descending aorta (DAo)\}. Here, we define the specialty by different anatomical structures, similar to specialty training in medical education. Given a phenotype set $\mathcal{P} = {p_1, p_2, ..., p_K}$, each agent evaluates:
$
\mathcal{V}_i = \{\phi_i(p_k, \mathcal{R}_i) | p_k \in \mathcal{P}\}
$
where $\phi_i(\cdot)$ represents agent $\alpha_i$'s phenotype valuation function incorporating statistical significance assessment, clinical relevance evaluation, population distribution analysis, and stability among the data. Each agent is able to use tools to extract features from statistical pools within the scope of prior knowledge for passing to the discussion. The collective phenotype valuations ${\mathcal{V}_i}$ form the fidelity of representation before the discussion. 

\texttt{Stage II: Associative Factor Discovery.} 
Then, agents examine a wide range of non-imaging factors $\mathcal{C} = {c_1, c_2, ..., c_M}$ through a similar two-level analysis:
For local analysis, each agent $\alpha_i$ evaluates factor effects within its domain:
$
\mathcal{E}_i = \{(p_n, c_m, \psi_i(p_n, c_m, \mathcal{V}_i)) | p_n \in \mathcal{P}, c_m \in \mathcal{C}\}
$
where $\psi_i(\cdot,\cdot)$ quantifies the phenotype-factor association strength. 
For global analysis, the coordinator agent aggregates local assessments $\{\mathcal{E}_i\}_{i=1}^N$ to establish cross-domain factor effects $\mathcal{E}_G = h(\{\mathcal{E}_i\}_{i=1}^N)$, where $h(\cdot)$ is a consensus aggregation function. The agents are set to reveal highly significant associations with a wide range of non-imaging factors, such as early-life factors, lifestyle and cognitive functions etc.

\texttt{Stage III: Collaborative Reasoning.} 
The final stage facilitates collaborative reasoning among agents to reach final consensus and report generation. This is inspired by multidisciplinary team (MDT) panel meetings in clinical decision making. We design a dynamic long-term memory mechanism for the agents in MESHAgents, enabling them to store, retrieve, and update memories like human experts. 
Agents can facilitate personalized analysis and diagnosis based on historical interactions. Meanwhile, agents interact through structured dialogues following protocols of evidence-based argumentation. The consensus formation follows an iterative process until convergence across all decision-making nodes (\textit{i.e.,} recommended phenotype, top associative factor, top diagnosis). 

\section{Experiments and Results}

\begin{figure*}[t]
\centering
\resizebox{0.7\columnwidth}{!}{%
\begin{minipage}{0.49\textwidth}
    \centering
    \resizebox{\textwidth}{!}{
    \begin{tabular}{l|cc}
    \toprule
    \multirow{2}{*}{Model} & \multicolumn{2}{c}{\texttt{Auto PheWAS}} \\
    \cmidrule(lr){2-3}
    & Dependency$\downarrow$ & Coverage$\uparrow$ \\
    \midrule
    \rowcolor{gray!15} \multicolumn{3}{c}{Independent LLMs (zero-shot CoT)} \\
    \midrule
    GPT-3.5~\cite{achiam2023gpt} & 0.642 & 0.524 \\
    GPT-4omini~\cite{achiam2023gpt} & \underline{0.356} & 0.767 \\
    Claude-3.5~\cite{anthropic2023claude} & 0.520 & 0.841 \\
    \midrule
    \rowcolor{gray!15} \multicolumn{3}{c}{Medical Multi-agent Frameworks} \\
    \midrule
    MedAgents~\cite{tang2023medagents} & 0.509 & 0.829 \\
    RareAgents~\cite{chen2024rareagents} & 0.363 & 0.629 \\
    \midrule
    w/o Mem. \& Tool & 0.492 & \underline{0.847} \\
    w/o Stage III & 0.385 & \textbf{0.871} \\
    \texttt{MESHAgents} & \textbf{0.350} & \textbf{0.871} \\
    \bottomrule
    \end{tabular}
    }
    \captionof{table}{Performance comparison of different Agent-based models on two metrics for Auto PheWAS. }
    \label{tab:result1}
\end{minipage}%
\hfill
\quad
\begin{minipage}{0.49\textwidth}
    \centering
    \includegraphics[width=\textwidth]{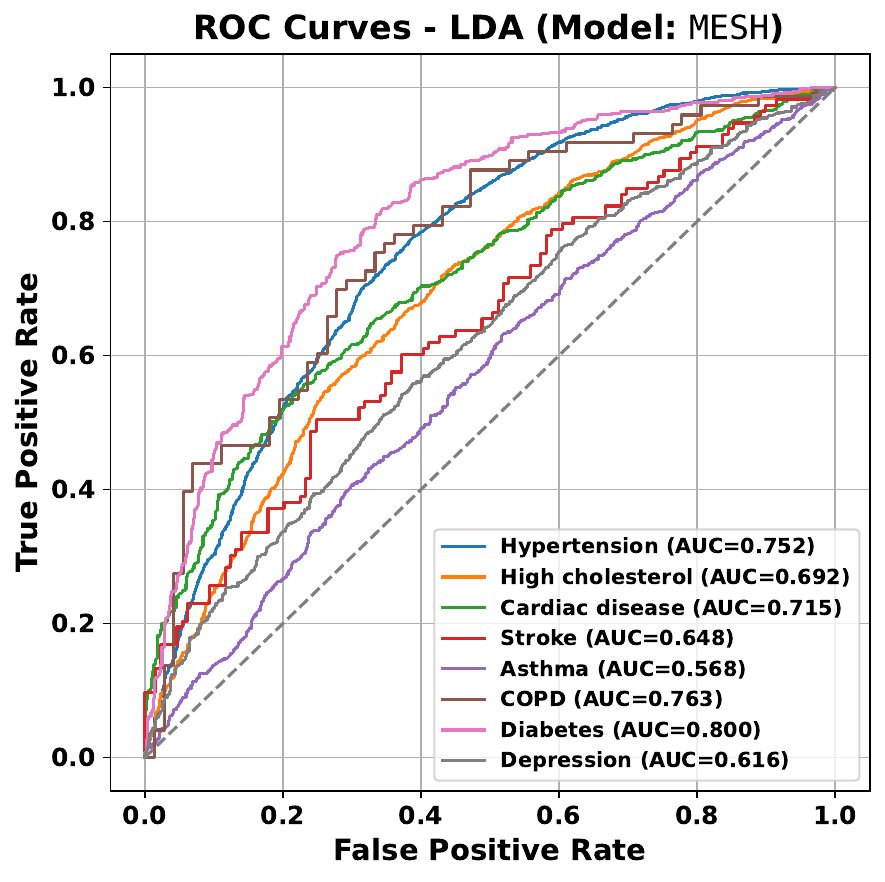}
    \captionof{figure}{ROC curves (LDA) for classification across cardiovascular conditions using discovered features.}
    \label{fig:clinical_outcome}
\end{minipage}
}
\end{figure*}

\noindent \textbf{Dataset and Setting.} 
We use UK Biobank data, which is available from UK Biobank via a standard application procedure\footnote{http://www.ukbiobank.ac.uk/register-apply}. Cardiac imaging phenotypes are derived from cardiac MR images using a deep learning-based pipeline~\cite{bai2020population} and released by the UK Biobank\footnote{https://biobank.ndph.ox.ac.uk/showcase/label.cgi?id=157}.
For generalizability and cross-validation, \texttt{MESHAgents} mine phenotypes and factors on a training set of 26,893 participants, and tested on the expanded dataset of 38,309 participants, which includes the initial cohort plus additional participants. Meanwhile, we perform hypothesis testing with existing expert knowledge in Fig.~\ref{fig:result1}, and performance comparison with automatic systems in Table~\ref{tab:result1} and Fig.~\ref{fig:clinical_outcome}. All experiments are performed on two Nvidia RTX 6000 GPUs (48 GB).

\noindent \textbf{Benchmarks.} 
(i) 
The LLMs we evaluate are the latest version of GPT-4o and GPT3.5~\cite{achiam2023gpt}. All models are set in a zero-shot Chain-of Thought (CoT)~\cite{wei2022chain} setting. 
(ii) For medical multi-agent frameworks: we select MedAgents~\cite{tang2023medagents} and RareAgents~\cite{chen2024rareagents} as the latest medical frameworks. For those LLMs that originally cannot complete the PheWAS task, we introduce the task definition into their agent initialization process for fair comparison.

\noindent \textbf{Evaluation Metrics.}
(i) In auto PHeWAS: We evaluate automatically discovered phenotypes through two complementary metrics:
\textit{Dependency:} Measures both phenotypic redundancy and hallucination tendencies. For phenotype set $\mathcal{P}$, calculated as $Q(\mathcal{P}) = [1 - \frac{2}{K(K-1)}\sum_{i=1}^{K-1}\sum_{j=i+1}^{K}|corr(p_i, p_j)|] \times \frac{|\mathcal{P}_{valid}|}{|\mathcal{P}|}$, where $|\mathcal{P}_{valid}|$ is the number of valid phenotypes in set $\mathcal{P}$. Lower values indicate greater phenotypic independence and fewer hallucinated features.
\textit{Coverage:} Assesses anatomical comprehensiveness across six anatomical structures (LV, RV, LA, RA, AAo, DAo) in~\cite{strocchi2020publicly,bai2020population} . We define coverage as $C(\mathcal{P}) = w_s \times \frac{|\Omega_{covered}|}{|\Omega|} + w_f \times \frac{|F_{covered}|}{|F_{total}|}$, where $\Omega_{covered}$ represents covered anatomical structures and $F_{covered}$ represents covered structure-function combinations. Higher coverage indicates global representation.
(ii) In disease diagnosis: We assess the clinical utility of the MESHAgents-discovered phenotype-factor associations $\mathcal{A}$, using these features to train classification models for nine disease types. We implement three classification algorithms (AdaBoost~\cite{freund1997decision}, LDA~\cite{fisher1936use}, and SVM~\cite{cortes1995support}) and report two evaluation metrics for classification performance: AUC, which measures overall discriminative ability, and recall, which quantifies the model's capacity to correctly identify positive disease cases. All results are obtained through five-fold cross-validation to ensure statistical reliability, and the practical value of automatically discovered phenotype-factor set.

\begin{table*}[t] 
\centering
\caption{Performance comparison between experts-selected and MESHAgentsdiscoved on disease classification tasks. (Green and bold values indicate where MESHAgents outperformed analyses using human-selected phenotypes. Summary: For SVM, MESHAgents showed $\overline{AUC}$ = $-0.013_{\pm0.076}$ and $\overline{Recall}$ = $-0.005_{\pm0.042}$; for LDA, $\overline{AUC}$ = $-0.004_{\pm0.010}$ and $\overline{Recall}$ = $\mathbf{+0.020_{\pm0.033}}$; for AdaBoost, $\overline{AUC}$ = $-0.014_{\pm0.036}$ and $\overline{Recall}$ = $-0.009_{\pm0.040}$.)}
\resizebox{\textwidth}{!}{
\begin{tabular}{lcccccccccccccccccc}
\toprule
 & \multicolumn{2}{c}{Hypertension} & \multicolumn{2}{c}{High cholesterol} & \multicolumn{2}{c}{Cardiac disease} & \multicolumn{2}{c}{PVD} & \multicolumn{2}{c}{Stroke} & \multicolumn{2}{c}{Asthma} & \multicolumn{2}{c}{COPD} & \multicolumn{2}{c}{Diabetes} & \multicolumn{2}{c}{Depression} \\
\cmidrule(lr){2-3} \cmidrule(lr){4-5} \cmidrule(lr){6-7} \cmidrule(lr){8-9} \cmidrule(lr){10-11} \cmidrule(lr){12-13} \cmidrule(lr){14-15} \cmidrule(lr){16-17} \cmidrule(lr){18-19}
& AUC & Recall & AUC & Recall & AUC & Recall & AUC & Recall & AUC & Recall & AUC & Recall & AUC & Recall & AUC & Recall & AUC & Recall \\
\midrule
\rowcolor{gray!15} \multicolumn{19}{c}{SVM} \\
\midrule
Experts & 0.739 & 0.708 & 0.645 & 0.672 & 0.671 & 0.548 & 0.316 & 0.462 & 0.398 & 0.805 & 0.549 & 0.691 & 0.561 & 0.521 & 0.730 & 0.670 & 0.573 & 0.667 \\
\texttt{MESHAgents}  & 0.726 & 0.706 & 0.635 & \cellcolor{green!15}\textbf{0.716} & 0.645 & 0.513 & 0.265 & 0.462 & \cellcolor{green!15}\textbf{0.619} & 0.779 & 0.549 & 0.676 & 0.549 & 0.438 & 0.678 & \cellcolor{green!15}\textbf{0.678} & 0.568 & \cellcolor{green!15}\textbf{0.679} \\
\midrule
\rowcolor{gray!15} \multicolumn{19}{c}{LDA} \\
\midrule
Experts & 0.753 & 0.687 & 0.692 & 0.655 & 0.722 & 0.614 & 0.479 & 0.308 & 0.644 & 0.628 & 0.570 & 0.567 & 0.760 & 0.630 & 0.801 & 0.732 & 0.619 & 0.625 \\
\texttt{MESHAgents}  & 0.752 & \cellcolor{green!15}\textbf{0.691} & 0.692 & \cellcolor{green!15}\textbf{0.660} & 0.715 & \cellcolor{green!15}\textbf{0.624} & \cellcolor{green!15}\textbf{0.521} & \cellcolor{green!15}\textbf{0.462} & \cellcolor{green!15}\textbf{0.648} & 0.619 & 0.568 & \cellcolor{green!15}\textbf{0.568} & \cellcolor{green!15}\textbf{0.763} & \cellcolor{green!15}\textbf{0.685} & 0.800 & 0.732 & 0.616 & 0.616 \\
\midrule
\rowcolor{gray!15} \multicolumn{19}{c}{AdaBoost} \\
\midrule
Experts & 0.747 & 0.705 & 0.676 & 0.670 & 0.701 & 0.609 & 0.487 & 0.538 & 0.587 & 0.602 & 0.548 & 0.542 & 0.697 & 0.644 & 0.775 & 0.716 & 0.595 & 0.611 \\
\texttt{MESHAgents}  & 0.743 & 0.700 & 0.673 & 0.658 & 0.689 & 0.601 & \cellcolor{green!15}\textbf{0.538} & 0.462 & \cellcolor{green!15}\textbf{0.619} & 0.566 & \cellcolor{green!15}\textbf{0.554} & \cellcolor{green!15}\textbf{0.552} & 0.649 & 0.644 & \cellcolor{green!15}\textbf{0.777} & \cellcolor{green!15}\textbf{0.724} & 0.580 & 0.598 \\
\bottomrule
\end{tabular}
\label{tab:result2}
}
\end{table*}

\noindent \textbf{Results.}
(i) Auto PheWAS: Independent LLMs generated unreliable phenotype sets with high dependency (GPT-3.5: 0.642), trade-off (GPT-4omini: 0.356 but 0.767), and limited coverage (GPT-3.5: 0.524), reflecting hallucination tendencies and catastrophic forgetting. Existing multi-agent frameworks showed improvement but exhibited polarization biases (RareAgents: 0.629), either over-focusing on specific imaging phenotypes or generating redundant phenotypes (MedAgents: 0.509). MESHAgents overcame these limitations through three key components: Firstly, evidence-augmented analysis with tools and memory significantly reduced hallucination; Second, the consensus mechanism promoted cross-checking while maintaining diversity; and the distributed expertise analysis prevented knowledge distortion. These components combined to achieve the best dependency score (0.350) and strong coverage (0.871) among all automated methods for PheWAS.

(ii) Disease Association Study: The comparison with human selected imaging phenotypes from literature~\cite{bai2020population,qiao2024personalised}. While MESHAgents do not fully match human patterns, it independently discovered significant relevant phenotypes without domain-specific knowledge. Fig.~\ref{fig:result1}(a) shows explainable overlap between MESHAgents-discovered and expert-selected parameters, with MESHAgents identifying alternative markers with diagnostic value. Importantly, MESHAgents achieves this automation while maintaining validity, as evidenced by its disease identification in Fig.~\ref{fig:result1}(b). This shows MESHAgents' ability to effectively balance automated exploration with clinical relevance, offering complementary advantages to traditional expert-driven phenotype association studies.

(iii) Diagnosis: Table~\ref{tab:result2} shows a comparison between MESHAgent-discovered and expert-selected imaging phenotypes in disease classification across three classifiers. Notably, MESHAgents shows comparable overall performance with mean AUC differences of only $-0.013_{\pm0.076}$ (SVM), $-0.004_{\pm0.010}$ (LDA), and $-0.014_{\pm0.036}$ (AdaBoost). For recall metrics—MESHAgent achieves $\overline{\text{Recall}}$ of $-0.005_{\pm0.042}$ (SVM), $+0.020_{\pm0.033}$ (LDA, outperforming human-selected features), and $-0.009_{\pm0.040}$ (AdaBoost). The superior recall stems from MESHAgents' ability to identify complementary phenotypes that better capture disease variance in linear discriminant space in Fig.~\ref{fig:clinical_outcome}. MESHAgents-discovered phenotypes achieved superior results in specific conditions like Stroke (AUC +0.221 in SVM), PVD (AUC +0.042 in LDA), and Diabetes. This suggests that MESHAgents' automatic phenotype selection captures clinically relevant patterns without requiring extensive domain knowledge. The classification results across Fig.~\ref{fig:clinical_outcome} further validate that MESHAgents-identified parameters maintain robust diagnostic value across different conditions, demonstrating the generalizability and scalability of our approach. These findings suggest that automatic phenotype discovery through multi-agent reasoning can effectively complement expert feature selection, potentially reducing subjective bias while maintaining clinical relevance in cardiovascular studies.

\section{Conclusion}

The framework could be further enhanced through the integration of vector embedding bases. For knowledge integration, an embedding library of medical imaging literature could be incorporated into our existing agent communication protocols, allowing phenotype-confounder evaluations to reference specific studies. In general, \texttt{MESHAgents} framework demonstrates advantages of multi-agent reasoning for medical imaging phenotype analysis compared to traditional approaches that rely on predefined criteria.

\noindent\textbf{Acknowledgements}
W.Z. is supported by the JADS programme and the UKRI Centre for Doctoral Training in AI for Healthcare (EP/S023283/1). M.Q. is supported by the EPSRC DeepGeM Grant (EP/W01842X/1) and the Dame Julia Higgins Postdoc Collaborative Research Fund. W.B. is supported by the EPSRC DeepGeM Grant (EP/W01842X/1), CVD-Net Programme Grant (EP/Z531297/1) and the BHF New Horizons Grant (NH/F/23/70013). S.N. is supported by the National Institutes of Health (R01-HL152256), the European Research Council (PREDICT-HF 864055), the British Heart Foundation (RG/20/4/34803), the EPSRC (EP/X012603/1 and EP/P01268X/1), the Technology Missions Fund under the EPSRC (EP/X03870X/1), and the Alan Turing Institute. P.M.M. acknowledges generous personal support from the Edmond J. Safra Foundation and Lily Safra, an NIHR Senior Investigator Award, Rosalind Franklin Institute, and the UK Dementia Research Institute, which is funded predominantly by the UKRI Medical Research Council. B.K. acknowledges the HPC resources provided by NHR@FAU under the NHR projects b143dc and b180dc. NHR funding is provided by federal and Bavarian state authorities. NHR@FAU hardware is partially funded by the DFG - 440719683. Additional support was  received by the ERC - project MIA-NORMAL 101083647,  DFG 513220538, 512819079, and by the state of Bavaria (HTA). This research was conducted using the UK Biobank Resource under Application Number 18545. We thank all UK Biobank participants and staff.

\noindent\textbf{Disclosure of Interests.}
The authors have no competing interests to declare that are relevant to the content of this article.


\newpage

\end{document}